%% file: AnomalyDetection.tex
\documentclass[conference]{IEEEtran}
\IEEEoverridecommandlockouts

\usepackage{amsmath,amssymb,amsfonts}
\usepackage{cite,cleveref,enumitem}
\usepackage{bbm,amsthm}
\usepackage{algorithm,algorithmic}
\usepackage{textcomp}
\usepackage{graphicx,xcolor}

\usepackage[english]{babel}

\usepackage{caption}
\usepackage{subcaption}	
\graphicspath{{Figures/}}
\usepackage{ifpdf}
\ifpdf
  \usepackage{epstopdf}
\fi

\usepackage{blindtext}

\def\BibTeX{{\rm B\kern-.05em{\sc i\kern-.025em b}\kern-.08em
    T\kern-.1667em\lower.7ex\hbox{E}\kern-.125emX}}

\crefname{figure}{fig.}{figs.}
\crefname{section}{sec.}{secs.}

\newcommand{\upi}{\pi_{\mathrm{upper}}}

\input{Supporting_Files/My_notations}

\begin{document}

\title{A Scalable Algorithm for Anomaly Detection via Learning-Based Controlled Sensing
}

\author{\IEEEauthorblockN{Geethu Joseph, M. Cenk Gursoy, and Pramod K. Varshney, \emph{Life Fellow, IEEE}}
\IEEEauthorblockA{\textit{Dept. of Electrical Engineering and Computer Science} \\
\textit{Syracuse University}\\
Syracuse, NY 13244, USA \\
Emails:\{gjoseph,mcgursoy,varshney\}@syr.edu.
}}


\maketitle

\begin{abstract}
We address the problem of sequentially selecting and observing processes from a given set to find the anomalies among them. The decision maker observes one process at a time and obtains a noisy binary indicator of whether or not the corresponding process is anomalous. In this setting, we develop an anomaly detection algorithm that chooses the process to be observed at a given time instant, decides when to stop taking observations, and makes a decision regarding the anomalous processes. The objective of the detection algorithm is to arrive at a decision with an accuracy exceeding a desired value while minimizing the delay in decision making. Our algorithm relies on a Markov decision process defined using the marginal probability of each process being normal or anomalous, conditioned on the observations. We implement the detection algorithm using the deep actor-critic reinforcement learning framework.  Unlike prior work on this topic that has exponential complexity in the number of processes, our algorithm has computational and memory requirements that are both polynomial in the number of processes. We demonstrate the efficacy of our algorithm using numerical experiments  by comparing it with the state-of-the-art methods.

\end{abstract}

\section{Introduction}
We consider the problem of observing a given set of processes to detect the anomalies among them via controlled sensing. Here, the decision maker does not observe all the processes at each time instant, but sequentially selects and observes one process at a time. The sequential control of the observation process  is referred to as controlled sensing. The challenge here is to devise a selection policy to sequentially choose the processes to be observed so that the decision is accurate and fast. This problem arises, for instance, in sensor networks used for remote health monitoring, structural health monitoring, etc~\cite{chung2006remote,bujnowski2013enhanced}. Such systems are equipped with different types of sensors to monitor different functionalities (or processes) of the system.  The sensors send their measurements to a common decision maker that identifies any potential system malfunction. These sensor measurements can be noisy due to faulty hardware or unreliable communication links. Therefore, to ensure the accuracy of the decision, we employ a sequential process selection strategy that observes the set of processes one at a time over multiple time instants  before the final decision is made. Further, the different processes can be dependent on each other, and therefore, observing one process also gives information about other dependent processes. Our goal is to derive a selection policy that accurately identifies the anomalous processes with minimum delay by exploiting the underlying statistical dependence among the processes.

A popular approach for solving the anomaly detection problem is to use the active hypothesis testing framework~\cite{zhong2019deep,joseph2020anomaly}. Here,  the decision maker defines a hypothesis corresponding to each of the possible states of the processes and computes the posterior probabilities over the hypothesis set using the observations. The decision maker continues to collect observations until the probability corresponding to one of the hypotheses exceeds the desired confidence level. This framework of active hypothesis testing was introduced by Chernoff in ~\cite{chernoff1959sequential}, and it was followed by several other studies in the literature~\cite{bessler1960theory,nitinawarat2013controlled,naghshvar2013active,huang2018active}. Recently, some researchers have combined the active hypothesis testing framework with deep learning algorithms to design data-driven anomaly detection algorithms~\cite{kartik2018policy,zhong2019deep,joseph2020anomaly,joseph2020anomaly2}. These algorithms learn from a training dataset and come with an added advantage of adaptability to the underlying statistical dependence among the processes. The state-of-the-art algorithms in this direction employ reinforcement learning (RL) algorithms such as Q-learning~\cite{kartik2018policy} and actor-critic~\cite{zhong2019deep,joseph2020anomaly}, and the active inference framework~\cite{joseph2020anomaly2}. However, the major drawback of this solution strategy is the heavy computational burden that arises due to the large number of hypotheses. Since each process can either be normal or anomalous, the number of hypotheses increases exponentially with the number of processes. Therefore, in this paper, we attempt to devise a learning-based controlled sensing framework for anomaly detection with polynomial complexity in the number of processes.  

The specific contributions of the paper are as follows: we first reformulate the problem of anomaly detection in terms of the marginal (not joint) probability of each process being normal or anomalous, conditioned on the observations. Consequently, the number of posterior probabilities computed by the algorithm at every time instant is linear in the number of processes. Based on these marginal posterior probabilities, we define the notion of a confidence level that is proportional to the decision accuracy, and a reward function that monotonically increases with the decision accuracy and decreases with the duration of the observation acquisition phase. These definitions allow us to reformulate the anomaly detection problem as a long-term average reward maximization of a Markov decision process (MDP). This problem is solved using a policy gradient RL algorithm called the actor-critic method, and the algorithm is implemented using deep neural networks. Using numerical results, we show that our algorithm is able to learn and adapt to the statistical dependence among the processes. Further, the polynomial complexity of the algorithms makes it scalable, and  hence, practically more useful.

\section{Anomaly Detection Problem}
We consider a set of $N$ processes where the state of each process is a binary random variable. The process state vector is denoted by $\vecs\in\{0,1\}^N$ whose $i\nth$ entry being $0$ and $1$ indicates that the $i\nth$ process is in the normal state and the anomalous state, respectively. We aim to detect the anomalous processes, which is equivalent to estimating the random binary vector $\vecs$. 

We estimate the process state vector $\vecs$ by selecting and observing one process at every time instant, and obtaining a state estimate of the corresponding process which has a finite probability of being erroneous. Let the process observed at time $k$ be $a(k)\in\lc1,2,\ldots,N\rc$ and the corresponding observation be $y_{a(k)}(k)\in\{0,1\}$. The uncertainty in the observation is modeled using the following probabilistic model:
\begin{equation}\label{eq:mesurement}
y_{a(k)}(k) = \begin{cases}
{ s_{a(k)}} & {\text{ with probability } 1-p,}\\
1-{ s_{a(k)}}& {\text{ with probability } p},
\end{cases}
\end{equation}
where $p\in[0,1]$ is called the flipping probability. Further, we assume that conditioned on the value of $\vecs$, the observations obtained across different time
instants are jointly (conditionally) independent, i.e., for any $k$, 
\begin{equation}
\bbP\ls \lc y_{i}(l), i=1,2,\ldots,N\rc_{l=1}^k\middle|\vecs\rs = \prod_{i=1}^N\prod_{l=1}^k\bbP\ls  y_{i}(l)\middle|s_i\rs.\label{eq:indep} 
\end{equation}
Therefore, the $i\nth$ process $\lc\vecy_i(k)\in\{0,1\}\rc_{k=1}^{\infty}$ is a sequence of independent and identically distributed (i.i.d.) binary random variables parameterized by $\vecs_i\in\{0,1\}$.

After each observation arrives, the decision maker computes an estimate of $\vecs$ along with the confidence in the estimate. The decision maker continues to observe the processes until the confidence exceeds the desired level denoted by $\upi\in(0,1)$. Therefore, we have two interrelated tasks: one, to develop an algorithm to estimate the process state vector and the associated confidence in the estimate; and two, to derive a policy that decides the process to be observed at each time instant and the criterion to stop collecting observations.  We seek the estimation algorithm and the policy that jointly minimize the stopping time $K$ while maximizing the accuracy level. Here, the stopping time refers to the time instant at which the observation acquisition phase ends.  We next present our estimation algorithm and policy design.

\section{Estimation Algorithm}\label{sec:estimation}
In this section, we derive an algorithm to estimate the process state vector from the observations. We note that the observations depend on the selection policy, and the policy design, in turn, depends on the estimation algorithm. Therefore, we first present the estimation algorithm and then derive a selection policy based on the estimation objectives in the next section.

To estimate the process state vector, we first compute the belief vector $\vecsigma(k)\in[0,1]^N$ at time $k$ whose $i\nth$ entry $\sigma_i(k)$ is the posterior probability that the $i\nth$ process is normal ($s_i=0$). Therefore, the probability that the $i\nth$ process is anomalous ($s_i=1$) is $1-\sigma_i(k)$. As each observation arrives, we recursively update the  belief vector as follows. 
\begin{align}
\sigma_i(k) &= \bbP\ls s_i= 0 \middle| \lc y_{a(l)}(l)\rc_{l=1}^{k} \rs\notag\notag\\
&=  \frac{\bbP\ls \lc y_{a(l)}(l)\rc_{l=1}^{k}\middle| s_i= 0 \rs\bbP\ls s_i=0\rs}{\bbP\ls\lc y_{a(l)}(l)\rc_{l=1}^{k} \rs}.\label{eq:posterior_1}
\end{align}
Here, we approximate the joint probability distribution by assuming that the observation $y_{a(k)}(k)$ is independent of the past observations $\lc y_{a(l)}(l)\rc_{l=1}^{k-1} $ conditioned on the process state $s_i$:
\begin{align}
\bbP\ls \lc y_{a(l)}(l)\rc_{l=1}^{k}\middle| s_i= 0 \rs\bbP\ls s_i=0\rs\notag\\
 &\hspace{-4.6cm}\approx \bbP\ls \lc y_{a(l)}(l)\rc_{l=1}^{k-1} \middle| s_i= 0 \rs\bbP \ls y_{a(k)}(k) \middle| s_i= 0 \rs\bbP\ls s_i=0\rs\notag\\
 &\hspace{-4.6cm}= \sigma_i(k-1)\bbP\ls\lc y_{a(l)}(l)\rc_{l=1}^{k-1} \rs\bbP \ls y_{a(k)}(k) \middle| s_i= 0 \rs.\label{eq:approx}
\end{align}
From \eqref{eq:indep}, the observation $y_{a(k)}(k)$ is independent of all other observations, conditioned on the value of $s_{a(k)}$. 
Therefore, the approximation is exact when $s_{a(k)}$ is a deterministic function of $s_i$. Some examples of such cases are $\bbP\ls s_{a(k)}=s_{i}\rs=1$, and $\bbP\ls s_{a(k)}=1-s_{i}\rs=1$. 
 
Substituting \eqref{eq:approx} into \eqref{eq:posterior_1}, we obtain
\begin{equation}
\sigma_i(k)=\frac{\sigma_i(k-1)\bbP \ls y_{a(k)}(k) \middle| s_i= 0 \rs}{\Sigma_i(k) },\label{eq:sigma_update1}
\end{equation}
Here, following the approximation in \eqref{eq:approx}, the normalization constant is
\begin{multline}
\Sigma_i(k) =\sigma_i(k-1)\bbP \ls y_{a(k)}(k) \middle| s_i= 0 \rs\\
+(1-\sigma_i(k-1))\bbP \ls y_{a(k)}(k) \middle| s_i= 1 \rs.\label{eq:sigma_update2}
\end{multline}
Further, the conditional probability $\bbP \ls y_{a(k)}(k) \middle| s_i= s \rs$ for $s=0,1$ is given by
\begin{align}
\bbP \ls y_{a(k)}(k) \middle| s_i= s \rs\notag \\
&\hspace{-2.5cm}=\displaystyle \sum_{s'=0,1}\bbP \ls y_{a(k)}(k) \middle| s_{a(k)}= s' \rs\bbP\ls s_{a(k)} =s'\middle| s_i=s\rs\notag\\
&\hspace{-2.5cm}=\displaystyle \sum_{s'=0,1}\bigg[ p^{\lv s'- y_{a(k)}(k)\rv }(1-p)^{\lv 1-s'- y_{a(k)}(k)\rv }\notag\\
&\hspace{1.5cm}\times \bbP\ls s_{a(k)} =s'\middle| s_i=s\rs\bigg ],\label{eq:sigma_update3}
\end{align}
which follows from \eqref{eq:mesurement}. Here, the term $\bbP\ls s_{j} =s'\middle| s_i=s\rs$ can be easily estimated from the training data\footnote{During the training phase, the true value of $\vecs$ is provided, but the optimal selection at each time instant is unknown.}
for every pair $(i,j)$.  Hence, \eqref{eq:sigma_update1}, \eqref{eq:sigma_update2}, and \eqref{eq:sigma_update3} give the recursive update of $\vecsigma(k)$.

We note that when $s_{a(k)}$ and $s_i$ are independent processes,
\begin{equation}
\bbP \ls y_{a(k)}(k) \middle| s_i= s \rs =  \bbP \ls y_{a(k)}(k) \rs \; s=0,1.
\end{equation}
Consequently, \eqref{eq:sigma_update1} reduces to $\sigma_i(k)=\sigma_i(k-1)$. This update is intuitive since an observation from process $s_{a(k)}$ does not change the probabilities associated with an independent process $s_i$. In other words, the recursive relation is exact when $s_i$ and $s_{a(k)}$ are either independent or $s_{a(k)}$ can be exactly determined from $s_i$.  We discuss this point in detail in \Cref{sec:simulations}.

Once $\vecsigma(k)$ is obtained,  the computation of the process state vector estimate denoted by $\hat{\vecs}(k)$ is straightforward:
\begin{equation}\label{eq:estimate}
\hat{s}_i = \begin{cases}
0 & \text{ if } \sigma_i(k)\geq 1-\sigma_i(k)\\
1 & \text{ if } \sigma_i(k)< 1-\sigma_i(k).
\end{cases}
\end{equation}

 Hence, the derivation of the estimation algorithm is complete. We next discuss the design of the selection policy. 

\section{Selection Policy}\label{sec:policy}
The design of the selection policy is a sequential decision making problem, and therefore, this problem can be formulated using the mathematical framework of an MDP. This formulation allows us to obtain the selection policy via reward maximization of the MDP using RL algorithms. In the following subsections, we define the MDP framework and describe the RL algorithm using the deep actor-critic method.

\subsection{Markov Decision Process}
An MDP has four components: state space, action space, state transition probabilities, reward function. In our case, these components are defined as follows:
\begin{itemize}[leftmargin=0.3cm]
\item  \emph{MDP state:} Our estimation algorithm is based on the belief vector $\vecsigma(k)$  that changes with time after each observation arrives. Therefore, we define $\vecsigma(k)\in[0,1]^N$ as the state of the MDP at time $k$. We note that the MDP state vector $\vecsigma(k)$ is different from the process state vector~$\vecs$.
\item \emph{Action:} The state of MDP depends on the observation which  in turn depends on the process selected by the policy. Thus, the action taken by the decision maker at time instant $k$ is the selected process $a(k)\in\{1,2,\ldots,N\}$. 
\item \emph{MDP State Transition:} For our problem, the MDP state $\vecsigma(k)$ at time $k$ is a deterministic function of the previous MDP state $\vecsigma(k-1)$, the action $a(k)$, and the observation $y_{a(k)}(k)$. Therefore, the MDP state transition is modeled by \eqref{eq:sigma_update1}, \eqref{eq:sigma_update2}, and \eqref{eq:sigma_update3}.
\item  \emph{Reward Function:} We seek a policy that maximizes the decision accuracy and minimizes the stopping time $K$. Here, we capture the decision accuracy using the uncertainty associated with each process conditioned on the observations. The uncertainty associated with the $i\nth$ process can be quantified using the entropy of its posterior distribution $\begin{bmatrix}
\sigma_i(k) & 1-\sigma_i(k)
\end{bmatrix}$.  Therefore, the instantaneous reward of the MDP is
\begin{equation}\label{eq:imm_reward}
r(k) = \sum_{i=1}^N H(\sigma_i(k-1))-H(\sigma_i(k)),
\end{equation}
where $H(x)=-x\log x-(1-x)\log (1-x)$ is the entropy. Then, the long term reward can be defined as the expected discounted reward of the MDP:
$
\bar{R}(k) = \sum_{l=k}^K\gamma^{l-k} r(l),$
where $\gamma\in(0,1)$ is the discount factor. The discounted reward formulation implies that a reward received $l$ time steps in the future is worth only $\gamma^l$ times what it would be worth if it were received immediately. Thus, this formulation minimizes the stopping time.
\end{itemize}

Having defined the MDP, we next describe the actor-critic RL algorithm that solves the long-term average reward maximization problem. 

\subsection{Deep Actor-Critic Algorithm}
The deep actor-critic algorithm is a deep learning-based RL technique that provides a sequential policy that maximizes the long-term expected discounted reward $\bar{R}(k) $ of a given MDP. The actor-critic framework  maximizes the discounted reward using two neural networks: actor and critic networks.  The actor learns a stochastic policy that maps the state of the MDP to a probability vector on the set of actions. The critic learns a function that evaluates the policy followed by the actor and gives feedback to the actor. Therefore, the two neural networks interact and adapt to each other. 

The components of the actor-critic algorithm are as follows:
\begin{itemize}[leftmargin=0cm]
\setlength\itemsep{1em}

\item [] \emph{Actor Network:} The actor takes the state of the MDP $\vecsigma(k-1)\in[0,1]^N$ as its input. Its output is the probability vector $\vecmu(\vecsigma(k-1);\alpha)\in[0,1]^N$  over the set of processes where $\alpha$ denotes the set of parameters of the actor neural network. The decision maker selects a process $a(k)\sim \vecmu(\vecsigma(k-1);\alpha)$, i.e., the $i\nth$ process is selected at time $k$ with probability equal to the $i\nth$ entry $\mu_i(\vecsigma(k-1);\alpha)$ of the actor output.
\item [] \emph{Reward Computation:} Once the process $a(k)$ is selected, the decision maker receives the corresponding observation $y_{a(k)}$, and the MDP state $\vecsigma(k-1)$ is updated to  $\vecsigma(k)$ as given by \eqref{eq:sigma_update1}. The decision maker also calculates the instantaneous reward $r(k)$ using  \eqref{eq:imm_reward}, and the reward value is fed to the critic along with the current and previous states of the MDP.
\item [] \emph{Critic Network:} The input to the critic at time $k$  is given by
\begin{equation*}
\vectheta(k) = \lb \vecsigma(k),\vecsigma(k-1),r(k)\rb \in[0,1]^N\times [0,1]^N\times \bbR.
\end{equation*}
The output of the critic is a scalar critique $\delta(\vectheta(k);\beta)$ where $\beta$ denotes the set of parameters of the critic neural network. This critique is computed based on the value function $V(\vecsigma(k))$ of the current MDP state as defined below:
\begin{equation*}
V^{\mu}(\vecsigma) = \expect{a(k)\sim\mu}{\bar{R}(k)\middle|\vecsigma(k)=\vecsigma}.
\end{equation*}
We note that $V^{\mu}(\vecsigma)$ is the expected average future reward when the MDP starts at state $\vecsigma$ and follows the policy $\mu(\cdot;\theta)$ thereafter. In other words, $V^{\mu}(\vecsigma)$ indicates the long term desirability of the MDP being in state $\vecsigma$. The scalar critique takes the form of a temporal difference (TD) error $\delta(\vectheta(k);\beta)$
\begin{equation}\label{eq:temporalerror}
\delta(\vectheta(k);\beta) = r(k)+\gamma \hat{V}(\vecsigma(k))-\hat{V}(\vecsigma(k-1)),
\end{equation}
where $\hat{V}$ is the value function estimate learned by the critic. A positive TD error indicates that the probability of choosing the current action should be increased for the future, and a negative TD error suggests that the probability of choosing $a(k)$ should be decreased.
\item [] \emph{Learning Actor Parameters:} The goal of the actor is to choose a policy such that the value function is maximized which in turn maximizes the expected average future reward. Therefore, the actor updates its parameter set $\alpha$ using the gradient descent step by moving in the direction in which the value function is maximized. The update equation for the actor parameters is given by 
\begin{equation}\label{eq:policy_gradient}
\alpha = \alpha^-+\delta(\vectheta(k);\beta)\nabla_{\alpha}[\log\mu_{a(k)}(\vecsigma(k-1);\alpha)],
\end{equation}
where $\alpha^-$ is the estimate of the network obtained in the previous time instant~\cite[Chapter 13]{sutton2018reinforcement}. 
\item [] \emph{Learning Critic Parameters:} The critic chooses its parameters such that it learns the estimate $\hat{V}(\cdot)$ of the state value function $V(\cdot)$ accurately. Therefore, the critic updates its parameter set $\beta$ by minimizing the square of the TD error $\delta^2(\vectheta(k);\beta)$.
\item [] \emph{Termination criterion:} The actor-critic algorithm continues to collect observations until the confidence level on the decision exceeds the desired level $\upi$. We define the confidence level on $\hat{s}_i$ as $\max\{\sigma_i(k),1-\sigma_i(k)\}$. Therefore, the stopping criterion is as follows:
\begin{equation}\label{eq:stopping}
\underset{{i=1,2,\ldots,N}}{\min} \max\{\sigma_i(k),1-\sigma_i(k)\} > \upi.
\end{equation}
\end{itemize}
The above components completely describe the actor-critic algorithm, and we next summarize the overall algorithm and discuss its complexity.

\section{Overall Algorithm}\label{sec:overall}
Combing the estimation algorithm in \Cref{sec:estimation} and the deep actor-critic method in \Cref{sec:policy}, we obtain our anomaly detection algorithm. The decision maker collects observations using the selection policy obtained using the actor-critic algorithm until the stopping criterion given in \eqref{eq:stopping} is satisfied. After the actor-critic algorithm terminates, the decision maker computes $\hat{\vecs}$ using \eqref{eq:estimate}. We present the pseudo-code of the overall procedure in \Cref{alg:ActorCritic}.

\begin{algorithm}[t]
\caption{\strut Actor-critic RL for anomaly detection}
\label{alg:ActorCritic}
\begin{algorithmic}[1]
\REQUIRE Discount rate $\gamma\in(0,1)$, Upper threshold on confidence $\upi\in(0.5,1)$

\ENSURE $\alpha,\beta$ with random weights, $\vecsigma(0)$ with the prior on each process (can be  learned from the training data)

\FOR {Episode index $= 1,2,\ldots$}
\STATE Time index $k =1$
\REPEAT 
\STATE Choose a process $a(k)\sim\mu(\vecsigma(k-1),\alpha)$ 
\STATE Receive observation $\vecy_{a(k)}(k)$
\STATE Compute $\vecsigma(k)$ using \eqref{eq:sigma_update1} - \eqref{eq:sigma_update3}
\STATE Compute instantaneous reward $r(k)$ using \eqref{eq:imm_reward}
\STATE Update the actor neural network using \eqref{eq:policy_gradient}
\STATE Update the critic neural network by minimizing the temporal error $\delta$ in \eqref{eq:temporalerror} with respect to $\beta$
\STATE Increase time index $k=k+1$
\UNTIL {\eqref{eq:stopping} is satisfied}
\STATE Declare the estimate $\hat{\vecs}$ using \eqref{eq:estimate}
\ENDFOR
\end{algorithmic}
\end{algorithm}

The computational complexity of our algorithm is determined by the size of the neural networks, the update of the posterior belief vector given by \eqref{eq:sigma_update1}-\eqref{eq:sigma_update3}, and the reward computation given by \eqref{eq:imm_reward}. Since all of them have linear complexity in the number of processes $N$, the overall computational complexity of our algorithm is polynomial in $N$. Also, the sizes of all the variables involved in the algorithm are linear in $N$ except for the pairwise conditional probability $\bbP\ls s_i\middle| s_j\rs$ for $i,j=1,2,\ldots,N$. Therefore, the memory requirement of the algorithm is $\calO(N^2)$. Hence, our algorithm possesses polynomial complexity, unlike the anomaly detection algorithms in \cite{joseph2020anomaly,joseph2020anomaly2} that have exponential complexity in $N$. Consequently, our algorithm is more applicable in practical settings.

It is straightforward to extend our algorithm to the case in which the decision maker chooses $n$ processes at a time. In that case, the output layer of the actor has $\binom{N}{n}$ neurons, and we need to update $\vecsigma(k)\in[0,1]^N$ using  the conditional probabilities of the form $\bbP\ls s_{i_1}, s_{i_2},\ldots,s_{i_{n}}\middle| s_j\rs$, for $1<i_1<i_2<i_3<\ldots<N$ and $j=1,2,\ldots,N$. Therefore, the overall computational  complexity of the resulting algorithm is polynomial in $N$ and the memory requirement is $\calO(N^{n+1})$.

\begin{figure*}[hptb]
\begin{subfigure}{6cm}
\includegraphics[width= 7cm]{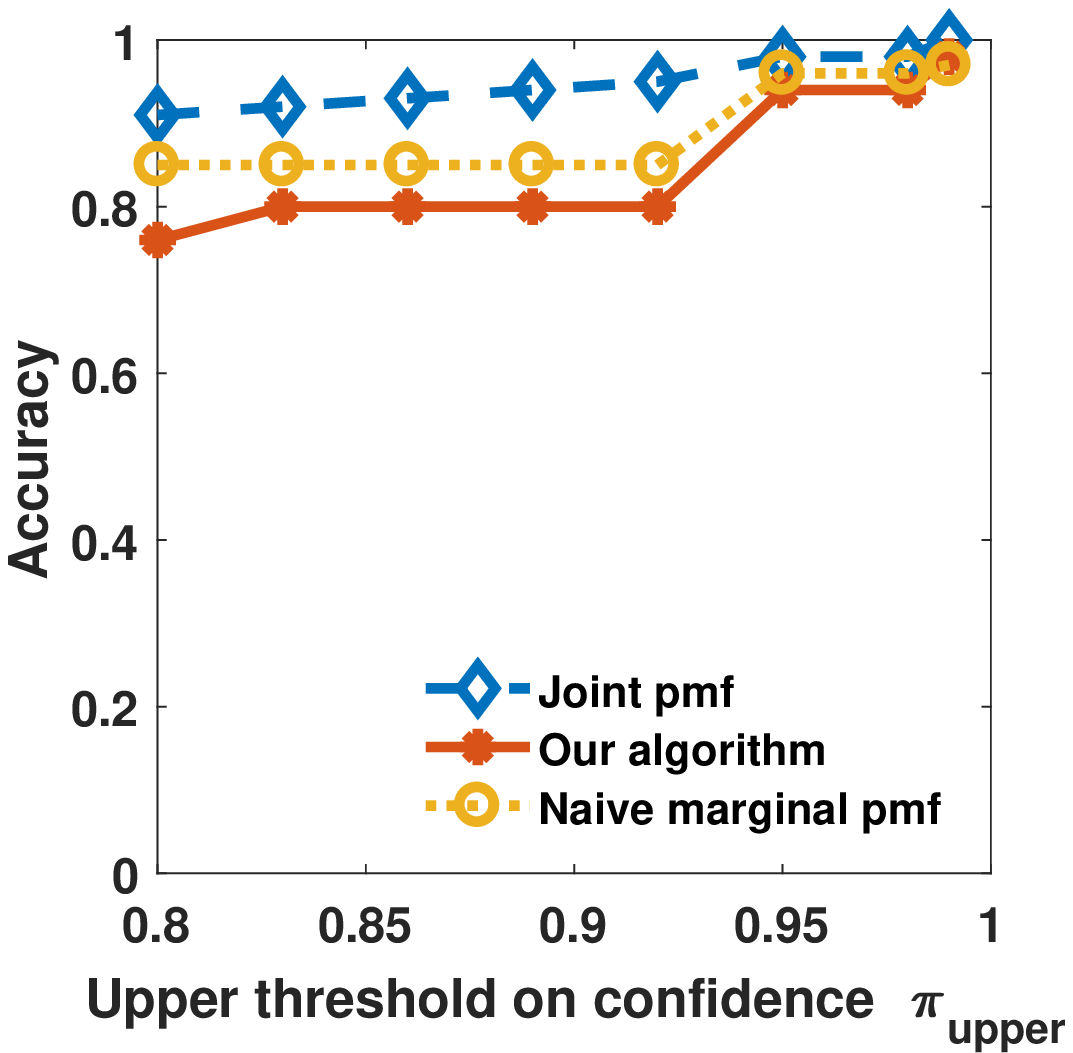}

\vspace{0.5cm}
\includegraphics[width= 7cm]{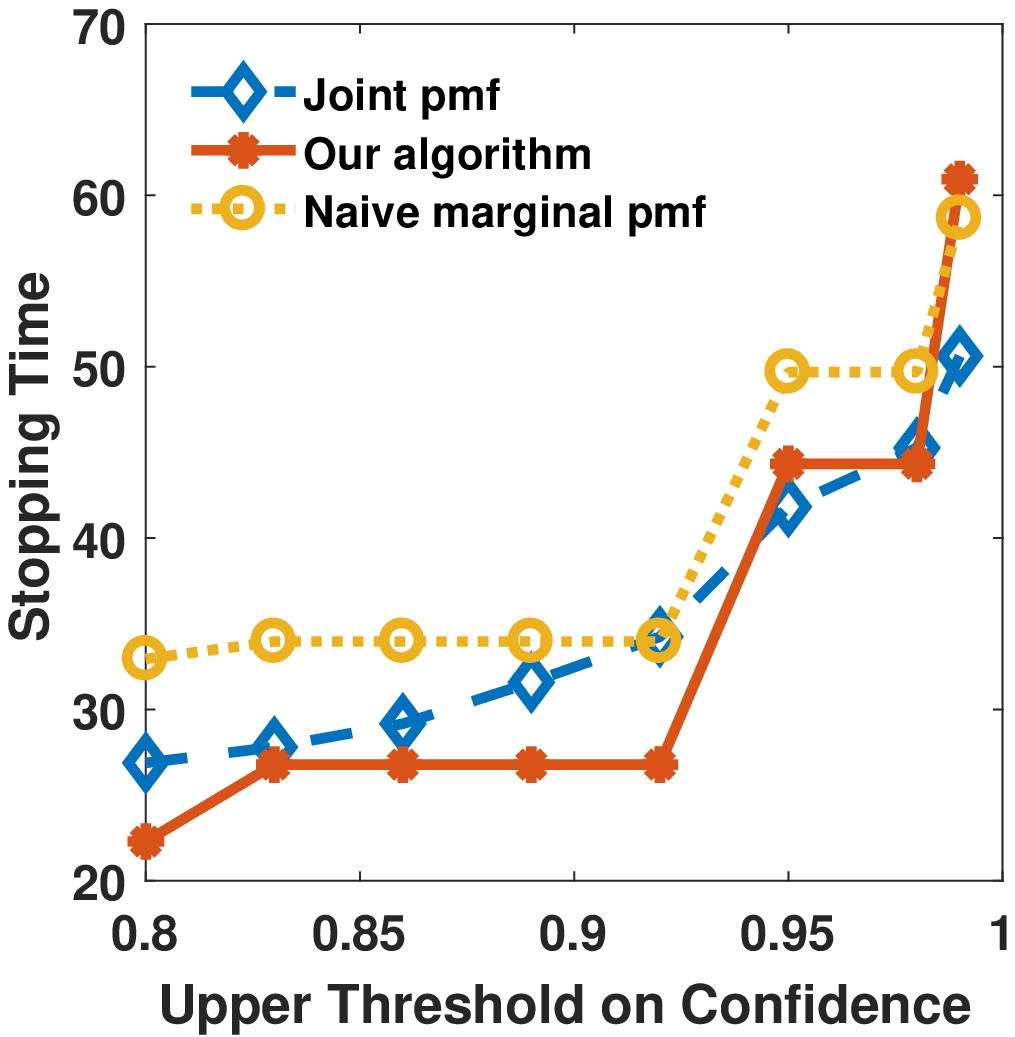}

\caption{Correlation coefficient $\rho = 0$}
\end{subfigure}
\begin{subfigure}{6cm}

\includegraphics[width= 7cm]{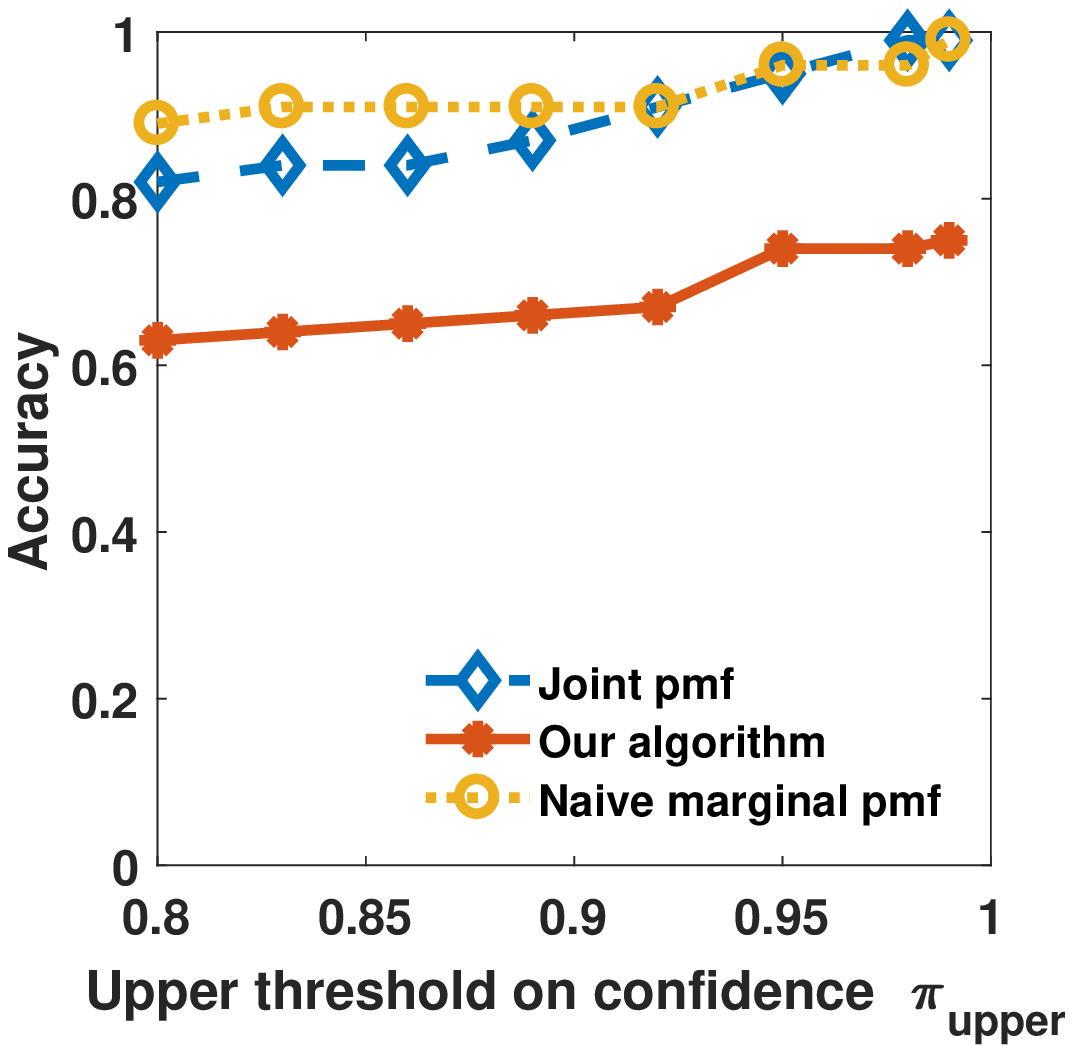}

\vspace{0.5cm}
\includegraphics[width= 7cm]{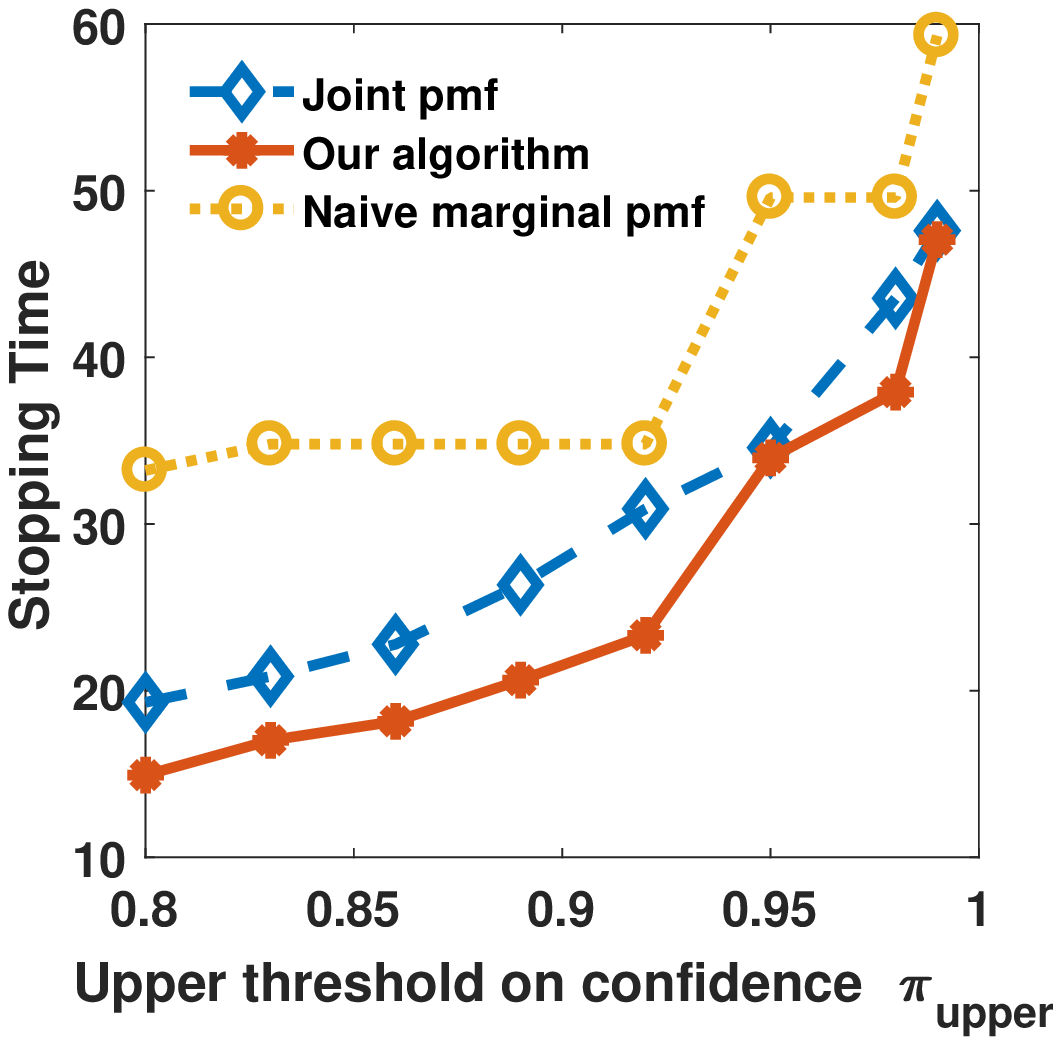}
\caption{Correlation coefficient $\rho = 0.6$}
\end{subfigure}
\begin{subfigure}{6cm}
\includegraphics[width= 7cm]{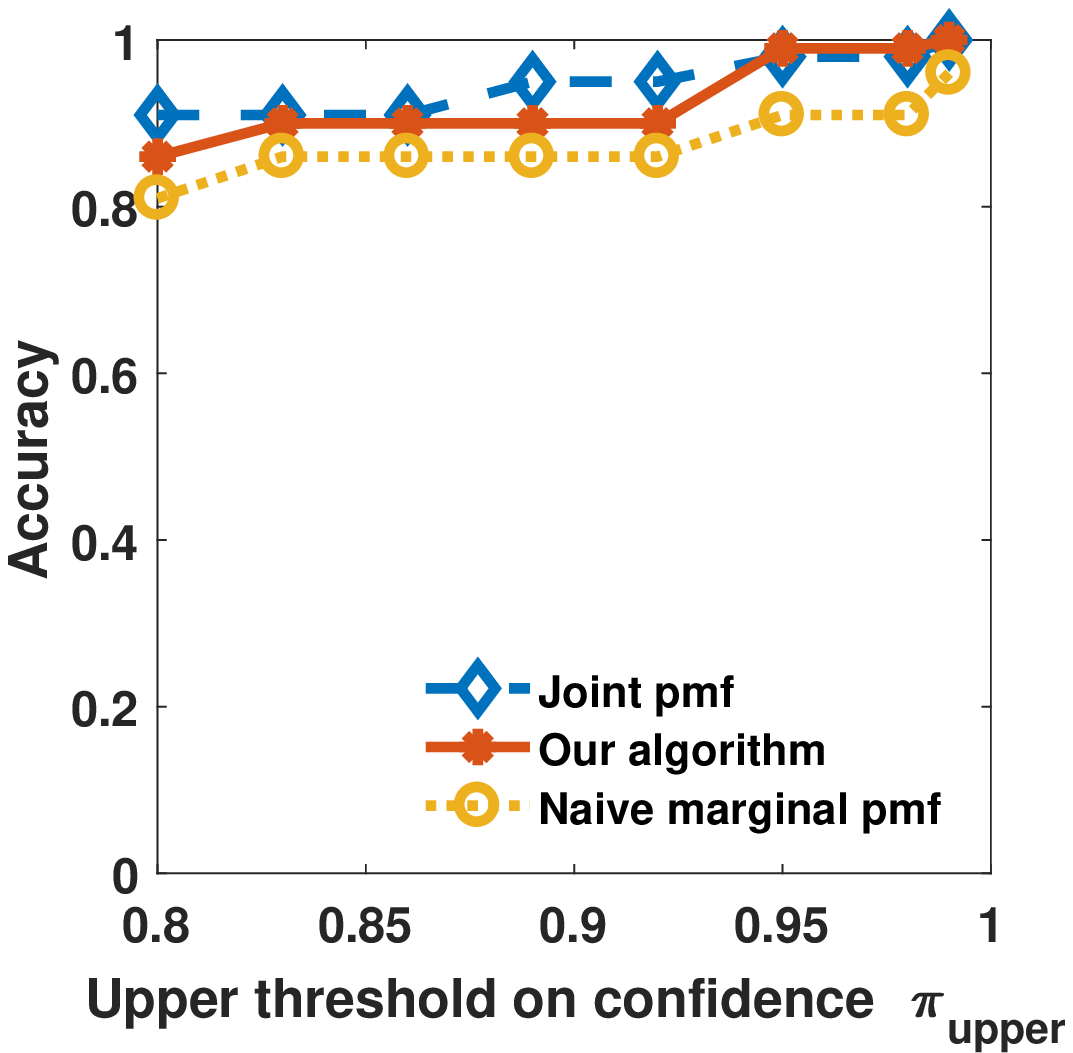}

\vspace{0.5cm}
\includegraphics[width= 7cm]{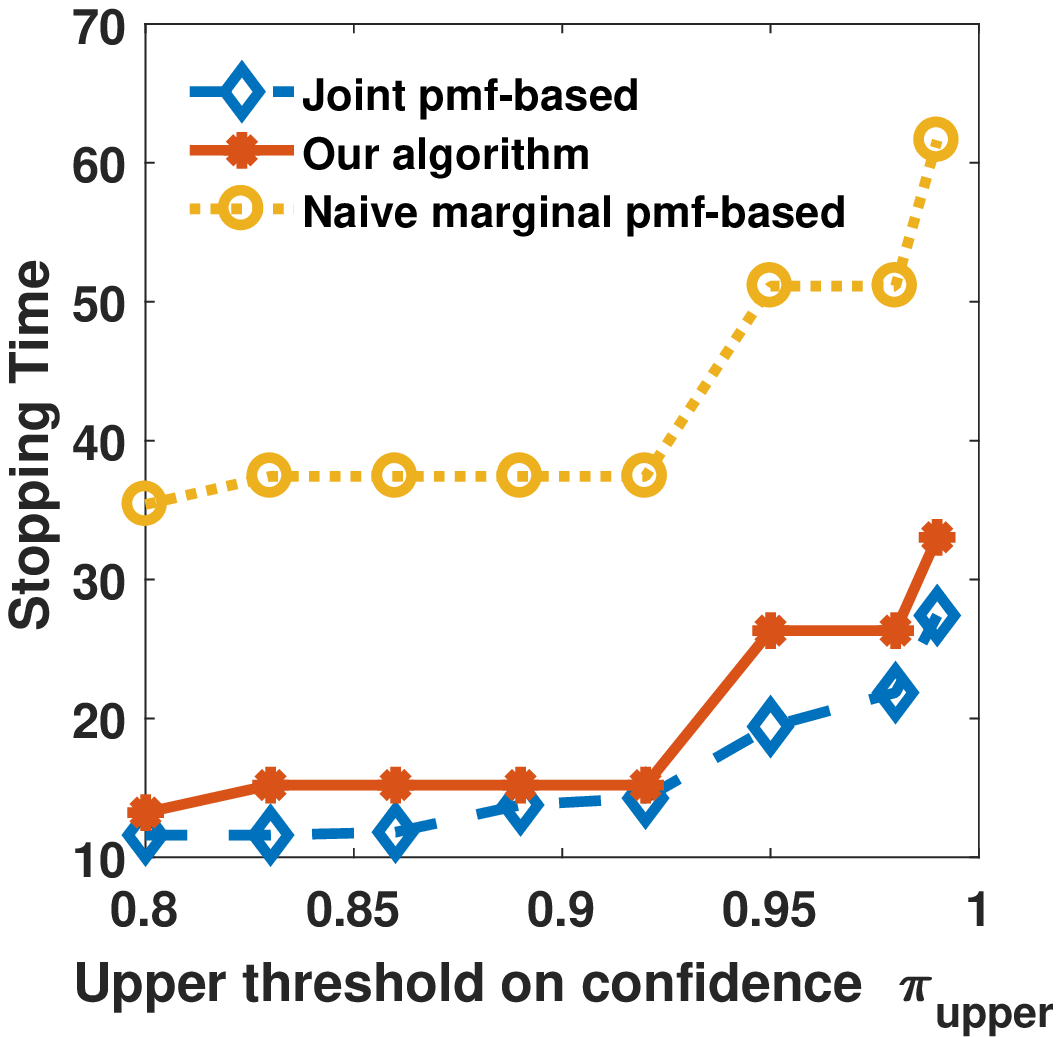}
\caption{Correlation coefficient $\rho = 1$}
\end{subfigure}
\caption{Performances of the three different deep actor-critic algorithms as a function of  $\upi$.}
\label{fig:ThresVs}
\end{figure*}

\begin{figure}[t]
\begin{center}

\begin{subfigure}{8cm}
\includegraphics[width= 7cm]{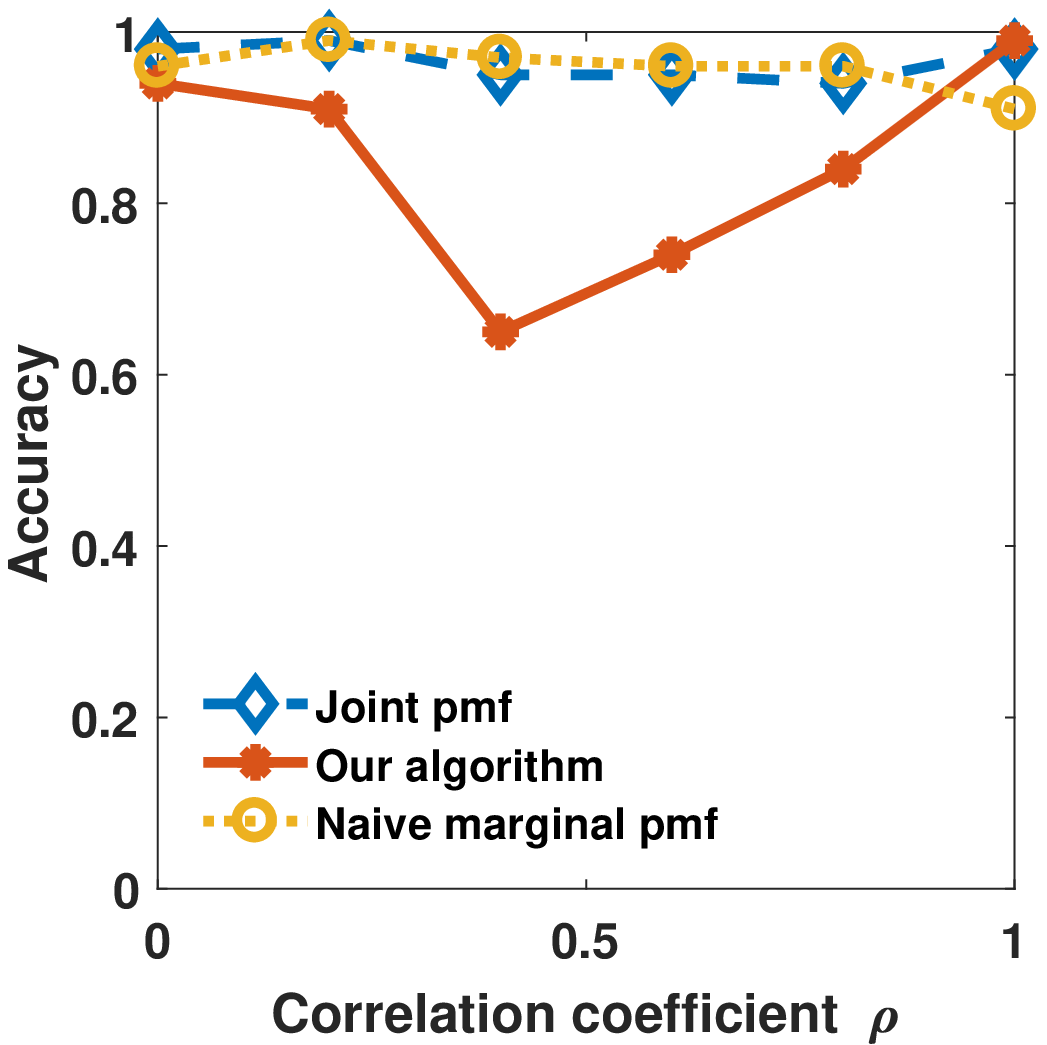} 

\includegraphics[width= 7cm]{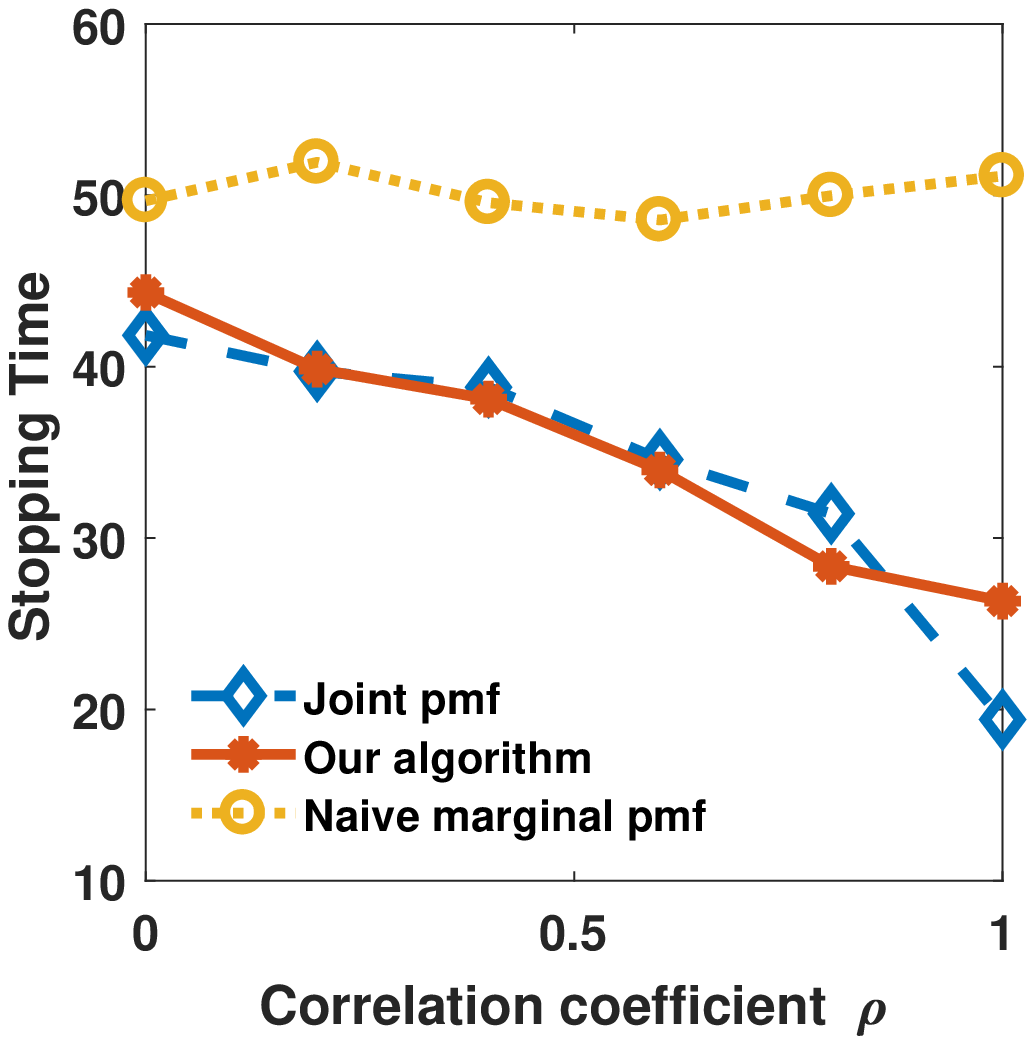}
\end{subfigure}

\end{center}
\caption{Performances of the three different deep actor-critic algorithms with $\rho$ as a function of  $\upi = 0.95$.}
\label{fig:RhoVs}
\end{figure}

\section{Simulation Results}\label{sec:simulations}
In this section, we empirically study the detection performance of our algorithm. We use two metrics for the performance evaluation: accuracy (the fraction of times the algorithm correctly identifies all the anomalous processes) and stopping time.

\subsection{Simulation Setup}
Our simulation setup is as described below:

\begin{itemize}[leftmargin=0cm]
\setlength\itemsep{1em}

\item [] \emph{Processes and Their Statistical Dependence:} We consider five processes $N=5$ and assume that the probability of each process being normal is $q=0.8$. Here, the first and second processes ($s_1$ and $s_2$) are statistically dependent, and the third and fourth processes ($s_3$ and $s_4$) are also statistically dependent. These pairs of processes are independent of each other and independent of the fifth process ($s_5$). The dependence is captured using the correlation coefficient $\rho\in[0,1]$ that is common to both process pairs:
 \begin{align*}
 \bbP\ls s_1=s_2=0\rs &= \bbP\ls s_3=s_4=0\rs = q^2+\rho q(1-q)\\
 \bbP\ls s_1\neq s_2\rs &=  \bbP\ls s_3\neq s_4\rs =(1-\rho) q(1-q)
 \end{align*}
 Also, we assume that the flipping probability $p=0.2$.
 
\item [] \emph{Implementation of Our Algorithm:} We implement the actor and critic neural networks with three layers and the ReLU activation function between consecutive layers. The output layer of the actor layer is normalized to ensure that $\mu(\cdot)$ is a probability vector over the set of processes. The parameters of the neural networks are updated using the Adam Optimizer, and we set the learning rates of the actor and the critic as $5\times 10^{-4}$, and $5\times 10^{-3}$, respectively. Also, we set the discount factor $\gamma = 0.9$.

\item [] \emph{Competing Algorithms:} We compare the performance of our algorithm with two other deep actor-critic-based algorithms:
\begin{enumerate}[leftmargin=0.3cm]
\setlength\itemsep{0.1em}
\item \emph{Joint probability mass function (pmf)-based algorithm:} This algorithm refers to the state-of-the-art method for anomaly detection problem presented in \cite{joseph2020anomaly}. The algorithm is based on the joint posterior probabilities of all the entries of $\vecs\in[0,1]^N$. Since $\vecs$ can take $2^N$ possible values, the complexity of this algorithm is $2^N$. However, the joint probabilities help the algorithm to learn all possible statistical dependencies among the process. 
\item \emph{Naive marginal pmf-based algorithm:} We also compare our algorithm with a naive method that also relies on the marginal probabilities $\vecsigma\in[0,1]^N$. This algorithm is identical to our algorithm except that at every time instant, this method only updates the entry of $\sigma_{a(k)}(k)$ of $\vecsigma(k)$ corresponding to the selected process $a(k)$. In other words, this method ignores the possible statistical dependence of the observation $y_{a(k)}(k)$ on the processes other than $a(k)$. Hence, the computational complexity of this algorithm is also $\calO(N)$. We note that unlike our algorithm, this algorithm does not use any approximation, and therefore, its updates are always exact.  

\end{enumerate} 
Our algorithm is a compromise between the above two algorithms and relies on marginal probabilities $\vecsigma$ while accounting for the possible statistical dependence among the processes. 
\end{itemize}
\subsection{Discussion of Results}
Our results are summarized in \Cref{fig:ThresVs,fig:RhoVs} and the key inferences from them are as follows:

\begin{itemize}[leftmargin=0.3cm]
\item The accuracy and the stopping time of all the algorithms increase with $\upi$. This trend is expected due to the fact that as $\upi$ increases, the decision maker requires more observations to satisfy the higher desired confidence level.
\item The accuracy of our algorithm is comparable to the other two algorithms when $\rho=0$ and $\rho=1$. The accuracy degrades as $\rho$ is close to 0.5. This behavior is because our algorithm uses approximate marginal probabilities to compute the confidence level whereas the other two algorithms use exact values. This approximation in \eqref{eq:approx} is exact when $\rho=0$ and $\rho=1$. As $\rho$ approaches $0.5$, the approximation error increases, and the accuracy decreases. 
\item The stopping times of the three algorithms are similar when $\rho=0$. This is because when $\rho=0$, all the processes are independent. Therefore, the updates of our algorithm are exact. The naive marginal pmf-based algorithm also offers good performance as there is no underlying statistical structure among the processes.
\item The stopping times of our algorithm and the joint pmf-based algorithm improve with $\rho$. As $\rho$ increases, the processes become more correlated, and therefore, an observation corresponding to one process has more information about the other correlated processes. However, the naive marginal pmf-based algorithm ignores this correlation and handles the observations corresponding to the different processes independently. Therefore, the stopping time is insensitive to $\rho$. Consequently, the difference between the stopping times of the naive marginal pmf-based algorithm and the other two algorithms increases as $\rho$ increases.
\end{itemize}

Further, from our experiments, we notice that the average runtime per per process selection decision for the joint pmf-based algorithm, our algorithm, and the naive marginal pmf-based algorithm are 3.2~ms,  2.88~ms, 2.89~ms, respectively. This observation is in agreement with our complexity analysis in \Cref{sec:overall} which implies that the joint pmf-based algorithm is computationally heavier compared to the other two algorithms. We also recall that the difference between the runtimes of the joint pmf based algorithm and our algorithm grows with $N$.

Thus, we conclude that our algorithm combines the best of two worlds by benefiting from the statistical dependence among the processes (similar to the joint pmf-based algorithm) and offering low-complexity (similar to the naive marginal pmf-based algorithm). 

\section{Conclusion}
We presented a low-complexity algorithm to detect the anomalous processes among a set of binary processes by observing a single process at a time. The sequential process selection problem was formulated using a Markov decision process whose reward is defined using the entropy of the marginal probabilities of the processes. The optimal process selection policy was obtained via the deep actor-critic algorithm that maximizes the long-term average reward of the MDP. Using numerical results, we established that our algorithm learns and adapts to the underlying statistical dependence among the processes while operating with low complexity. This algorithm relies on approximate marginal probabilities which can lead to  performance deterioration when the approximation error is large. A theoretical analysis that quantifies the approximation error is an interesting direction for future work.
\bibliographystyle{IEEEtran}
\bibliography{Supporting_Files/IEEEabrv,Supporting_Files/bibJournalList,Supporting_Files/AnomalyDetection}

\end{document}

%% file: Supporting_Files/My_notations.tex
\crefname{app}{Appendix}{Appendices}
\crefname{cor}{Corollary}{Corollary}
\crefname{prop}{Proposition}{Proposition}
\crefname{lemma}{Lemma}{Lemma}
\crefname{defn}{Definition}{Definition}
\crefname{conj}{Conjecture}{Conjecture}
\crefname{exam}{Example}{Example}
\crefname{supp}{Supplemental Section}{Supplemental Section}

\newcommand{\bs}{\boldsymbol}
\newcommand{\bb}{\mathbb}
\newcommand{\mcal}{\mathcal}

\newcommand{\lb}{\left(}
\newcommand{\rb}{\right)}
\newcommand{\ls}{\left[}
\newcommand{\rs}{\right]}
\newcommand{\lc}{\left\{}
\newcommand{\rc}{\right\}}

\newcommand{\lv}{\left\vert}
\newcommand{\rv}{\right\vert}

\newcommand{\LRV}[1]{{\left\vert\kern-0.25ex\left\vert\kern-0.25ex\left\vert #1 \right\vert\kern-0.25ex\right\vert\kern-0.25ex\right\vert}}

\newcommand{\expect}[2]{\bb{E}_{#1}\lc#2\rc}

\newcommand{\nth}{^\mathsf{th}}

\newcommand{\bbP}{\bb{P}}

\newcommand{\bbR}{\bb{R}}

\newcommand{\calO}{\mcal{O}}

\newcommand{\vecs}{\bs{s}}

\newcommand{\vecy}{\bs{y}}

\newcommand{\vecmu}{\bs{\mu}}

\newcommand{\vectheta}{\bs{\theta}}

\newcommand{\vecsigma}{\bs{\sigma}}